\documentclass[12pt]{article}

\usepackage{prime_arxiv}

\usepackage[utf8]{inputenc} 
\usepackage[T1]{fontenc}    
\usepackage[hyphens,spaces,obeyspaces]{url}            
\usepackage[colorlinks,allcolors=blue]{hyperref}       
\usepackage{booktabs}       
\usepackage{amsfonts}       
\usepackage{amsmath}
\usepackage{amssymb}
\usepackage{nicefrac}       
\usepackage{microtype}      
\usepackage[numbers]{natbib}
\usepackage{doi}

\usepackage{changepage}
\usepackage{enumitem}  

\usepackage{graphicx}
\usepackage{caption}
\usepackage{subcaption}
\usepackage{wrapfig} 

\graphicspath{ {./images/} } 

\title{MineSegSAT: An automated system to evaluate mining disturbed area extents from Sentinel-2 imagery}

\author{
    Ezra MacDonald \\
    University of Victoria \\
    Victoria, BC \\
    \texttt{macdonaldezra@gmail.com}
    \And 
    Derek Jacoby \\
    University of Victoria \\
    Victoria, BC \\
    \texttt{derekja@gmail.com}
    \And
    Yvonne Coady \\
    University of Victoria \\
    Victoria, BC \\
    \texttt{ycoady@gmail.com} \\
}

\hypersetup{
    pdftitle={MineSegSAT: An automated system to evaluate mining disturbed area extents from Sentinel-2 imagery},
    pdfsubject={ai.CS},
    pdfauthor={Ezra MacDonald, Derek Jacoby, Yvonne Coady},
    pdfkeywords={remote sensing, semantic segmentation, environmental monitoring, mineral extraction},
}

\begin{document}
\maketitle

\begin{abstract}
    Assessing the environmental impact of the mineral extraction industry plays a critical role in understanding and mitigating the ecological consequences of extractive activities. This paper presents MineSegSAT, a model that presents a novel approach to predicting environmentally impacted areas of mineral extraction sites using the SegFormer deep learning segmentation architecture trained on Sentinel-2 data. The data was collected from non-overlapping regions over Western Canada in 2021 containing areas of land that have been environmentally impacted by mining activities that were identified from high-resolution satellite imagery in 2021 \cite{tang_global_2023}. The SegFormer architecture, a state-of-the-art semantic segmentation framework, is employed to leverage its advanced spatial understanding capabilities for accurate land cover classification. We investigate the efficacy of loss functions including Dice, Tversky, and Lovasz loss respectively. The trained model was utilized for inference over the test region in the ensuing year to identify potential areas of expansion or contraction over these same periods. The Sentinel-2 data is made available on Amazon Web Services through a collaboration with Earth Daily Analytics which provides corrected and tiled analytics-ready data on the AWS platform. The model and ongoing API to access the data on AWS allow the creation of an automated tool to monitor the extent of disturbed areas surrounding known mining sites to ensure compliance with their environmental impact goals.
\end{abstract}

\keywords{remote sensing \and semantic segmentation \and environmental monitoring \and mineral extraction}

\section{Introduction}
This paper introduces MineSegSAT, a deep learning image segmentation model using the SegFormer architecture that is designed to segment areas of land that have been environmentally impacted by mineral extraction operations. The mining industry has seen a considerable expansion in recent years \cite{maus_update_2022} driven by growing demand for raw materials \cite{lenzen_implementing_2022} and demand trends indicate this growth will continue \cite{un_irp_global_2019}. While this sector is important for the industrialization of the global economy, mining sites can have adverse impacts on the immediate and near environment during mining operations and after closure. Identifying specifically environmentally impacted areas of land can benefit regulators and mining operations internally to ensure environmentally conscientious practices are upheld.

This project extracts Sentinel-2 tiles in Western Canadian provinces including British Columbia and Alberta that overlap with features of mining sites like waste rock dumps, pits, water ponds, tailing dams, heap leach pads and processing/milling infrastructure \cite{tang_global_2023}. The training, validation, and test datasets use 12 of the 13 bands captured from Sentinel-2, omitting Band 10 as it is typically used for atmospheric correction. 134 tiles were extracted from Sentinel-2 tiles, each spanning 7,680m$^{2}$ and their mining types were manually annotated based on overlapping OpenStreetMap views and the Canadian Minerals and Mining Map \cite{government_of_canada_natural_2016}. More than 85 percent of the tiles that are included in the dataset contain at least one mining pixel, though only roughly 4 percent of the total pixels included in the dataset contain environmentally impacted sections of land. This paper proposes a lightweight implementation of the SegFormer model architecture, a vision transformer-based deep learning model, which was introduced in 2021, by Xie, E. et. al \cite{xie_segformer_2021} with the minor modification that the most performant model replaces the cross entropy loss function used in the original work with the Tversky Loss function \cite{salehi_tversky_2017}.

\section{Related Work}

Deep learning has broad and significant applications in the context of remote sensing and environmental monitoring while Sentinel-2 data has proven to be useful for land cover and land use mapping and for improving the automation of environmental monitoring \cite{phiri_sentinel-2_2020}. In the context of monitoring mineral extraction operations, deep learning models and remote sensing data have been used to assess the significance of environmentally impacted areas \cite{gerassis_ai_2021} and to identify unregistered and illegal mining operations \cite{balaniuk_mining_2020} \cite{noauthor_learning_2019}.

The original transformer architecture proposed an encoder-decoder architecture using multi-head attention and feed-forward layers respectively to globally attend to information from a pair of input sequences \cite{vaswani_attention_2017}. The self-attention component in the architecture updates the value for each element in a sequence by aggregating information from the entire input. Variations of the original architecture that typically do not include cross-attention components detailed in the original architecture have since been applied to computer vision problems such as image classification and segmentation with considerable success. Both image classification and segmentation transformer-based models respectively have demonstrated state-of-the-art performance on benchmark data sets \cite{dosovitskiy_image_2020} \cite{zheng_rethinking_2021} \cite{liu_swin_2022} \cite{cheng_masked-attention_2022}. Transformer-based model architectures have similarly been applied to remote sensing segmentation and classification benchmarks with performance that rivals and in many cases improves on the performance of state-of-the-art model architectures \cite{li_deep_2022}.

In this paper, we explore using an implementation of the SegFormer model, a simple semantic segmentation model that uses a UNet architecture - unifying Transformer encoders with lightweight multilayer perception (MLP) decoders. This model is of particular interest since it uses lightweight multi-layer perception (MLP) layers in the decoder and outputs multi-scale features from the Transformer encoder layers. Since this model is outputting multi-scale features, it does not need to train positional embedding layers which can lead to decreased performance when the training and testing resolutions differ \cite{xie_segformer_2021}. In the context of remote sensing, this model has been used to extract information for water bodies \cite{yang_watersegformer_2023}, detect buildings using optical remote sensing images \cite{li_method_2023}, segment coastal wetlands from Sentinel-2 data \cite{lin_semantic_2023}, and for performing road segmentation \cite{ma_semi-supervised_2022}. 

\section{Data}

\begin{wrapfigure}{L}{0.2\textwidth}
    \centering
    \includegraphics[scale=0.4]{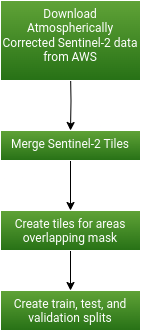}
\end{wrapfigure}

This model was trained using Sentinel-2 tiles, extracted from Amazon Web Services through a collaboration with Earth Daily Analytics which provides atmospherically corrected analytics-ready data on Amazon Web Services. The observation dates for the tiles used for training were from April 1st to September 1st in 2021 for the training period and then a comparison of tiles overlapping the test dataset in 2021 were compared from 2022 for inference and inspection. The tiles that were chosen have less than 1 percent cloud coverage and the tiles within the same coordinate reference system (CRS) were merged using the reverse painter’s algorithm. The intersecting ground truth masked data from \cite{tang_global_2023} was reprojected to the CRS of the merged Sentinel-2 tiles and then converted to a raster with 10m resolution to match the resolution of the input tiles.

Given that mining sites in Canada are subject to federal, provincial, and territorial environmental laws and are identified by organizations like Natural Resources Canada \cite{government_of_canada_natural_2016}, the dataset overwhelmingly includes tiles that include environmentally impacted areas of mineral extraction sites identified by Tang and Werner \cite{tang_global_2023}. From the mineral extraction sites that were identified in the original paper, Figure 1 shows the types of mineral extraction operations that could be confidently intersected with the ground truth masks used to train this model based on visual inspection with overlapping mapped areas using OpenStreetMap and \cite{government_of_canada_natural_2016}.

\begin{wrapfigure}{R}{0.3\textwidth}
    \vspace{-0.5cm}
    \centering
    \small
    \begin{tabular}{ |c|c|c|c| }
     \hline
      \textbf{Mining Operation} & \textbf{\# of Sites}\\
     \hline
     Coal & 40 \\ 
     Aggregate & 28 \\
     Precious Metals & 31 \\ 
     Mercury & 4 \\
     Uranium & 3 \\
     \hline
    \end{tabular}
    
    \captionof{figure}{Types of Mineral Extraction Operations Identified}
    \vspace{-1cm}
\end{wrapfigure}

Each mask and corresponding band file for each period were then split into tiles with dimensions 768x768, 384x384, and 128x128 pixels respective to resolutions of 10m, 20m, and 60m. Note that 4 percent of the pixels included in the dataset used for training, validation, and testing include environmentally impacted areas of mining sites and 86 percent of the 134 tiles include environmentally impacted sites. The dataset can be downloaded from Google Drive using {\href{https://drive.google.com/drive/folders/1FMruAwQeOB0T8BunxzBmjQI5R5uj6wAp?usp=sharing}{this link}}.

\section{Model}

The SegFormer model was introduced in 2021 as a lightweight transformer-based image segmentation model with an encoder-decoder or UNet architecture. Differing from the originally proposed Transformer architecture which requires encoding positional information for inputs, the SegFormer encoder layers use feed-forward layers to output multi-scale features, which avoids the need to interpolate positional codes which can lead to decreased performance when training and testing input resolutions differ \cite{xie_segformer_2021}. The model architecture avoids complex decoders, instead using multilayer perceptron (MLP) layers that aggregate information from each feature map generated by encoder layers. The model architecture used for this experiment is very similar to the B3 implementation-sized model proposed in the original SegFormer paper \cite{xie_segformer_2021} and differs in that the number of transformer layers used in the second of four layers in the encoder is 4 instead of the 3 used in the original model. This distinction was made as including the additional encoder layer proved to be beneficial in early model verification training iterations.

\begin{center}
    \centering
    \includegraphics[scale=0.4]{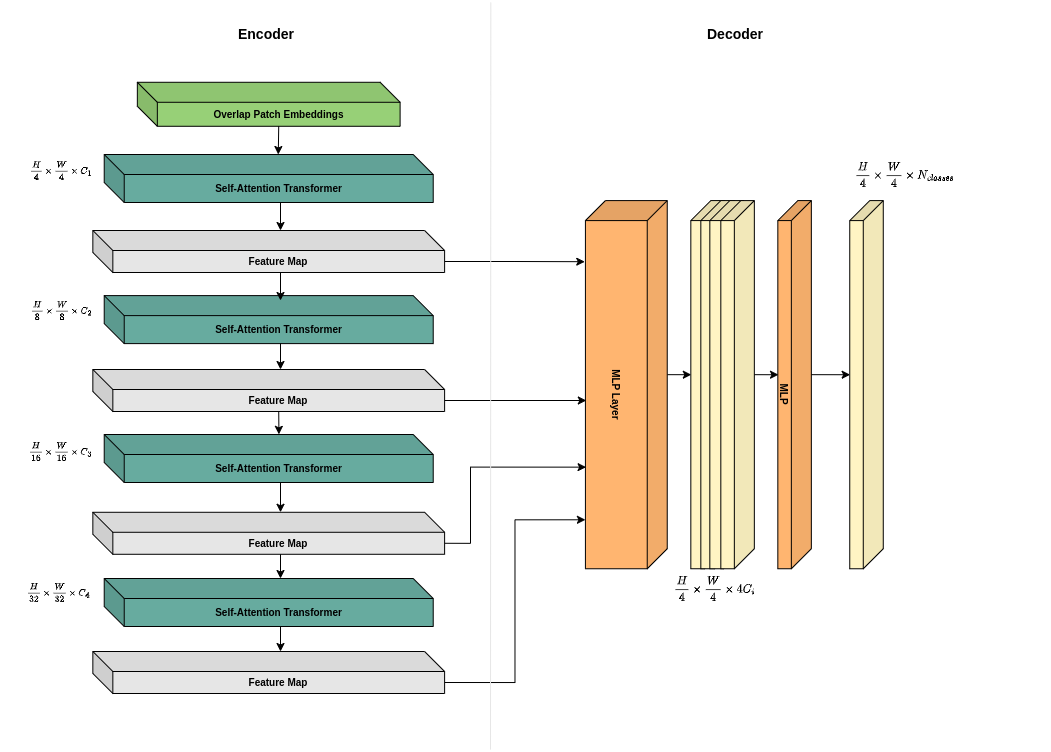}
    \captionof{figure}{SegFormer Model Architecture}
\end{center}

\section{Experiments}

Training of this model was performed using loss functions including Tversky Loss, Dice Loss, and Lovasz-Hinge Loss. For brevity, we focus on the results of the Tversky Loss function since it outperformed the other loss functions. Consider an image as a tensor $x \in [0, 1]^{L \times H \times W}$ with a corresponding segmentation mask $y \in \{ 0, 1 \}^{C \times H \times W}$, and let $\hat{y} \in \{ 0, 1 \}^{C \times H \times W}$, where $C = 1$ since we are using this loss function for binary segmentation. Now let $\alpha, \beta > 0$. Then the Tversky Loss denoted by $T$ corresponding to $\textbf{y}$ and $\hat{y}$ is given by

\begin{center}
    $T_{\alpha, \beta}(y, \hat{y}) = \dfrac{y\hat{y}}{y\hat{y} + \alpha (1 - y)\hat{y} + \beta y(1 - \hat{y}) + \delta }$
\end{center}

\noindent where $\delta > 0$ is a smoothing coefficient. Observe that setting $\alpha = \beta = 0.5$ gives us the Dice Loss and setting $\alpha = \beta = 1$ produces the Jaccard Loss. 

The metrics that were used for evaluating the efficacy of the predictions outputted by the model were the F1-Score, Precision, and Recall as stated below, where True Positives are given by $TP$; False Positives are given by $FP$; and False Negatives are given by $FN$.
\begin{align*}
    Precision &= \dfrac{TP}{TP + FP} \\
    Recall &= \dfrac{TP}{TP + FN} \\
    F1_{score} &= \dfrac{2 \times Precision \times Recall}{Precision + Recall}
\end{align*}

\noindent Each of these metrics are computed on a per-pixel basis for each image and the returned score is the average of the score for each image.
\begin{wrapfigure}{L}{0.4\textwidth}
    \centering
    \small
    \begin{tabular}{ |c|c|c|c| }
     \hline
      \textbf{Train Transforms} & \textbf{Probability}\\
     \hline
     Random Crop & 100\% \\ 
     Vertical Flip & 50\% \\ 
     Random Rotation & 50\% \\
     Horizontal Flipping & 50\% \\
     Channel Shuffle & 30\% \\
     \hline
    \end{tabular}
    \captionof{figure}{Transformations applied during training}
\end{wrapfigure}

Given that the band dimensions vary in their resolution, each of the respective 20m and 60m bands were up-sampled to the 10m resolution bands using inter-cubic interpolation. Before feeding data to the model during training, validation, and testing stages, data transformations were applied to each instance. The transforms applied during training were applied with a given probability detailed in Figure 3 on each input tile, while only random cropping was used for validation data, and patch merging for the test dataset. The model size that was used on the dataset was 512x512 and the batch size used during training and validation was 8. During the test stage, individual images were fed into the model. 

\begin{center}
    \includegraphics[scale=0.4]{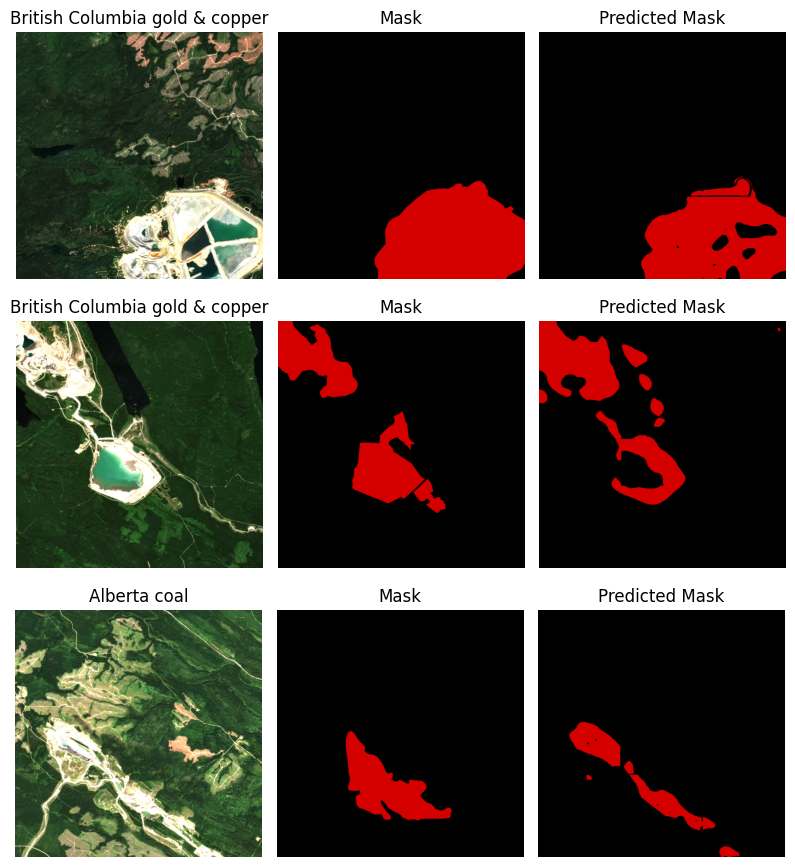}
    \captionof{figure}{Model predictions on test dataset}
\end{center}

The model was trained for 1,000 epochs with a learning rate of 0.0003 using a Cosine Annealing learning rate and was then fine-tuned for another 1,000 epochs with a starting learning rate of 0.0001. From the dataset, 70\% of samples were used for training while 15\% were used for validation and testing respectively. An additional experiment was performed to verify the results of the test dataset on the test region in the ensuing year, which did not result in additional environmentally impacted mining areas being identified. Code that accompanies this paper can be found \href{https://github.com/macdonaldezra/MineSegSAT}{here}. The results for the most performant model across the training, test, and validation datasets were as follows:

\begin{center}
\begin{tabular}{ |c|c|c|c| }
 \hline
  & \textbf{F1} & \textbf{Precision} & \textbf{Recall} \\
 \hline
 \textbf{train} & 0.5743 & 0.5514 & 0.5992 \\ 
 \textbf{val} & 0.7261 & 0.7216 & 0.5461 \\ 
 \textbf{test} & 0.5035 & 0.4890 & 0.5189 \\ 
 \hline
\end{tabular}
\end{center}

\section{Conclusion \& Future Work}

The ground truth data provided and models trained on this data are demonstrably capable of identifying the features identified in the original paper, though more work is required to assess the quality of the ground truth dataset and improve the performance of the model such that it might be useful for performing accurate environmental monitoring that can be reliably used in real-time. Correlating the identified mining areas with existing publicly identified above-ground or visibly environmentally impacted mining areas could be beneficial for better understanding the efficacy of the data and areas where the model is challenged in making accurate predictions.

From this sample dataset, the model struggles to identify characteristics like tailing ponds as seen in Figure 4 as well as built-up rock piles. Waterways like rivers, lakes, and potentially non-impacted water sources were included in the dataset including alpine and mountainous regions which may explain the model's challenge in identifying the respective features. Further training could be done that uses a larger dataset with additional mining sites and more generalized model pre-training in an effort to improve the robustness of the model. Data sources like Sentinel-1 data that include pertinent information about ground features and more data take periods over mining areas could be used in the training and evaluation datasets to provide a more enriched analysis of a given area. Further, other deep learning architectures should be trained on the dataset to establish a performance benchmark on the dataset and improve the understanding of areas of strength and weakness in predicting the masked regions.

A model trained based on these findings could present value for identifying and assessing multiple risks posed by mineral extraction operations by identifying anomalies in the size variation of tailing ponds, rock pile formations, open-pit mines, and processing/milling infrastructure which were identified in the ground truth masks. This model would thrive in a circumstance where it is monitoring known active or previously active mineral extraction sites that pose a potential threat to the environment. Given that this data is made available by government agencies \cite{government_of_canada_natural_2016}, implementing a meaningful deep learning-based monitoring service of this nature is not only possible but could significantly benefit the environmental monitoring of mineral extraction sites.

\bibliographystyle{unsrt}
\bibliography{main}

\end{document}